\documentclass[letterpaper, 10 pt, conference]{ieeeconf}  % Comment this line out if you need a4paper

\IEEEoverridecommandlockouts                              % This command is only needed if 
\usepackage{color}
\usepackage{graphics} % for pdf, bitmapped graphics files
\usepackage{epsfig} % for postscript graphics files
\usepackage{mathptmx} % assumes new font selection scheme installed
\usepackage{times} % assumes new font selection scheme installed
\usepackage{amsmath} % assumes amsmath package installed
\usepackage{amssymb}  % assumes amsmath package installed

\newcommand{\mwst}{\omega}
\newcommand{\mobs}{x}
\newcommand{\mact}{a}

\newcommand{\mWst}{\Omega}
\newcommand{\mObs}{X}
\newcommand{\mAct}{A}

\usepackage{graphicx}
\usepackage{subcaption}
\title{\LARGE \bf
An information-theoretic on-line update principle \\for  perception-action coupling
}

\author{Zhen Peng $^{1,2,3,\dagger}$, Tim Genewein$^{4}$, Felix Leibfried$^{5}$, Daniel A. Braun$^{6}$% <-this % stops a space
\thanks{$\ast$This study was supported by the ERC, Starting Grant BRISC 678082 and DFG, Emmy Noether grant BR4164/1-1.}% <-this % stops a space
\thanks{$^1$ Max Planck Institute for Biological Cybernetics, T\"{u}bingen, Germany}%
\thanks{$^2$ Max Planck Institute for Intelligent Systems, T\"{u}bingen, Germany}%
\thanks{$^3$ Graduate Training Centre of Neuroscience, T\"{u}bingen, Germany}%
\thanks{$^4$ Bosch Center for Artificial Intelligence, Robert Bosch GmbH, Renningen, Germany}%
\thanks{$^5$ PROWLER.io, Cambridge, United Kingdom}%
\thanks{$^6$ Institute for Neural Information Processing, University of Ulm, Ulm, Germany}%
\thanks{$\dagger$ Correspondence: Zhen Peng, Max Planck Institute for Biological Cybernetics, Spemannstr. 38, 72076 T\"{u}bingen, Germany
{\tt\small zhen.peng@tuebingen.mpg.de}}%
}

\begin{document}

\maketitle
\thispagestyle{empty}
\pagestyle{empty}

%%%%%%%%%%%%%%%%%%%%%%%%%%%%%%%%%%%%%%%%%%%%%%%%%%%%%%%%%%%%%%%%%%%%%%%%%%%%%%%%
\begin{abstract}

Inspired by findings of sensorimotor coupling in humans and animals, there has recently been a growing interest in the interaction between action and perception in robotic systems \cite{Bogh2016}. Here we consider perception and action as two serial information channels with limited information-processing capacity. We follow \cite{Genewein2015} and formulate a constrained optimization problem that maximizes utility under limited information-processing capacity in the two channels. As a solution we obtain an optimal perceptual channel and an optimal action channel that are coupled such that perceptual information is optimized with respect to downstream processing in the action module. The main novelty of this study is that we propose an online optimization procedure to find bounded-optimal perception and action channels in parameterized serial perception-action systems. In particular, we implement the perceptual channel as a multi-layer neural network and the action channel as a multinomial distribution. We illustrate our method in a NAO robot simulator with a simplified cup lifting task.

%Under limitation of computational resources, a bounded-rational decision-maker interacts with the environment by trading-off utility maximization against information-processing costs. In the information-theoretic framework the bounded-optimal action is given by self-consistent solutions which can be computed through the Blahut-Arimote algorithm. However, such iterative method can become computationally costly due to a exhaustively evaluation of the utility function for every possible outcome and it cannot be applied straightforwardly to continues problems. 

\end{abstract}

%%%%%%%%%%%%%%%%%%%%%%%%%%%%%%%%%%%%%%%%%%%%%%%%%%%%%%%%%%%%%%%%%%%%%%%%%%%%%%%%

\section{INTRODUCTION}

In robotic systems perception and action have often been studied in isolation in the past without an overarching principle of how to put the two processes together. Yet, there is compelling evidence that the two processes are interdependent in humans and animals \cite{Noe2004}.  
In the robotics literature, the direct coupling between action and perception has been especially emphasized in behavior-based robotics \cite{Arkin1998} and by proponents of embodied cognition \cite{Pfeifer2006}, but more recently also approaches applying machine learning to sensorimotor processing have focused on the  interactive nature of perception---see \cite{Bogh2016} for a review. The main insight of interactive perception is that sensory processing can be enhanced when manipulating or interacting with  the environment. This can be achieved by creating novel signals through movement \cite{Fitzpatrick2002, Hoof2014} or by exploiting action-perception regularities that are generated when the same action is performed repeatedly in the same environment \cite{Gupta2012, Chang2012}. For example, object segmentation could be improved when separating different objects by movement, or some object properties like inertia or weight could be estimated through interaction. In such cases action directly subserves the perceptual process. In other cases, however, interactive perception has the primary objective to achieve a manipulation goal. Defining the objective is therefore critical in determining what kind of sensorimotor coupling can arise.

The formal framework that deals with adaptive systems optimizing arbitrary objective functions under uncertainty is decision theory. A rational agent has to decide which action to take from the action set according to the desirability of the action quantified by a utility function. A fundamental problem of such perfect rationality models \cite{Ramsey1931, VonNeumann1944, Savage1954} is that they ignore computational costs that arise when searching for the maximum utility action. As such costs can be prohibitive, decision-making with limited information-processing resources has recently been studied extensively in psychology, economics, cognitive science, computer science, and artificial intelligence research \cite{Gigerenzer1999,  Kahneman2003, Howes2009, Russell1995, Russell2002, Lewis2014}. 
%\subsection{A Free Energy Principle for Bounded Rationality}
In the following we argue that such resource limitations are crucial for the emergence of sensorimotor coupling.

\subsection{An Information-Theoretic Principle for Bounded Rational Decision-Making with Context-dependence}

In this study, we use an information-theoretic model of bounded rational decision-making \cite{Braun2011,  Braun2014, Ortega2012, Ortega2013, Ortega2014b}. In a decision-making task with context, an agent is presented with a world-state $\mwst$ and has to find an optimal action $\mact^*_\mwst$ from a set of admissible actions. The desirability of the action under a particular world-state is quantified by the utility function $U$. The objective of the decision-maker is to maximize the utility depending on the context:

\begin{equation}  
 \mact^*_\mwst = \underset{\mact}{\operatorname{argmax}}{\,\,U(\mwst, \mact)}
 \end{equation}  

For an agent with limited computational resources that has to react within a certain time-limit, searching for the best action can potentially become intractable, especially when the number of possible actions is enormous. Thus, a bounded rational agent tries to find a good enough yet tractable solution. In multiple contexts, bounded rational decision-making  requires to compute multiple strategies under limited computational resources which can be expressed as a set of probability distributions $p(\mact|\mwst)$ over actions given the different world-states. Mathematically, this informational cost can be measured in terms of an ``information distance", namely the Kullback-Leibler divergence $D_{KL}(p(\mact|\mwst)||p_0(\mact))$ from a prior behavior $p_0(\mact)$ to the posterior strategies $p(\mact|\mwst)$. This information-processing cost can be motivated on axiomatic grounds \cite{Ortega2010, Ortega2013} and has been used previously in the robotics and control literature \cite{Peters2010, Theodorou2010, Todorov2006, Kappen2005}. An upper bound of the Kullback-Leibler divergence $D_{KL}(p(\mact|\mwst)||p_0(\mact)) \leqslant B$ with  $B\geqslant0$ constrains the decision-maker to spend a maximum number of bits $B$ to adapt its behavior. The resulting optimization problem can be formalized as 

\begin{multline}
 p^*(\mact|\mwst) = \underset{p(\mact|\mwst)}{\operatorname{argmax}}{\sum_{\mact}p(\mact|\mwst)U(\mwst, \mact)}\\
  - \frac{1}{\beta}D_{KL}(p(\mact|\mwst)||p_0(\mact)) 
\end{multline}
where the inverse temperature $\beta > 0$  governs the trade-off between expected utility and information cost. For $\beta \rightarrow \infty$ classic decision theory is recovered, whereas for $\beta \rightarrow 0$ the decision-maker has no access to computational resources at all and thus acts according to the prior. When optimizing the expected free energy over all possible contexts, it can readily be shown that the optimal prior $p_0(\mact)$ is given by the marginal distribution $p_0(\mact) = p(\mact) = \sum\limits_{\mwst}p(\mact|\mwst)p(\mwst)$ [see Section 2.1.1 in \cite{Tishby1999} or \cite{Csiszar1984}]. Plugging in the marginal $p(\mact)$ as the optimal prior $p_0^*(\mact)$ yields the following variational principle for bounded rational decision-making

\begin{equation}  
 p^*(\mact|\mwst) = \underset{p(\mact|\mwst)}{\operatorname{argmax}}{\mathbf{E}_{p(\mwst,\mact)}[U(\mwst, \mact)] - \frac{1}{\beta}I(\mWst;\mAct)}
 \label{eq:rd}
 \end{equation}  
where $I(\mwst;\mact)$ is the mutual information between action and world-states and measures the reduction of uncertainty about the action $a$ after observing $\mwst$ or vice versa. This problem formulation is commonly known as the rate-distortion problem from information theory \cite{Shannon1959}. The solution of (\ref{eq:rd}) is given by a set of two self-consistent equations:
 
 \begin{eqnarray}
p^*(\mact|\mwst) &=& \frac{1}{Z(\mwst)}p(\mact)\mathrm{exp}(\beta U(\mwst, \mact)) \label{eq:BA_1} \\
p(\mact) &=& \sum\limits_{\mwst}p^*(\mact|\mwst)p(\mwst) \label{eq:BA_2}
\end{eqnarray} 
where $Z(\mwst) = \sum\limits_{\mact'}p(\mact')\mathrm{exp}(\beta U(\mwst, \mact'))$ is the partition sum. In practice, the solution can be computed using the Blahut-Arimoto algorithm \cite{Blahut1972, Arimoto1972, Yeung2008} by starting with an initial distribution $p_{init}(\mact)$ and then iterating through both equations (\ref{eq:BA_1}) and (\ref{eq:BA_2}) until the distributions converge. The iteration is guaranteed to converge to a global optimum with the prerequisite that $p_{init}(\mact)$ has the same support as $p(\mact)$ \cite{Tishby1999, Csiszar1984, Cover1991}. 

%In the limit-case $\beta \rightarrow \infty$ where transformation costs are ignored, $p(\mact|\mobs)$ is equal to the perfectly rational policy for each value of $\omega$ \emph{independent} of any of the other policies and $p(\mact)$ becomes a mixture of these solutions. Importantly, high values of the mutual information term in Equation~\ref{Eq:RateDistortionVariational} will not lead to a penalization, which means that actions $\mact$ can be very informative about the observation $\mobs$. The behavior of an actor with infinite computational resources will thus in general be very observation-specific.
%
%In the case $\beta \rightarrow 0$ the mutual information between actions and observations is minimized to $I(\mact;\mobs)=0$, leading to $p(\mact|\mobs)=p(\mact)~\forall \mobs$, the maximal abstraction where all $\mobs$ elicit the same response. Within this limitation the actor will, however, still emit actions that maximize the expected utility $\sum_{\mact,\mobs} p(\mact) U(\mact,\mobs)$.
%
%For values of the rationality parameter $\beta$ in between these limit-cases, that is $0 < \beta < \infty$, the bounded-rational actor trades off \emph{observation-specific} actions that lead to a higher expected utility for particular observations at the cost of a high mutual information term, against \emph{abstract} actions that yield a ``good'' expected utility for many observations and lead to a lower mutual information term.

\subsection{An Information-Theoretic Principle for Perception-Action Coupling}

We follow the work of \cite{Genewein2015}, where the authors extend the rate-distortion framework to systems with multiple information-processing nodes. The serial perception-action system consists of two stages: a perceptual stage $p(\mobs|\mwst)$ that maps world-states $\mwst$ to observations $\mobs$ and an action stage $p(\mact|\mobs)$ that maps observations $\mobs$ to actions $\mact$.  The three random variables for world-state, observation and action form a serial chain of two channels, which is expressed by the graphical model $ \mWst \rightarrow \mObs \rightarrow \mAct$, and implies the following conditional independence 

\begin{equation}  
p(\mwst, \mobs, \mact) = p(\mwst)p(\mobs|\mwst)p(\mact|\mobs)
\end{equation}  

We assume that the utility function depends only on the world-state and the action $U(\mwst, \mact)$, the internal percept $\mobs$ does not influence the utility. The information processing price in the perceptual channel $\beta_{1}$ can be different from the price of information processing in the action channel $\beta_{2}$. Formally, we set up the following variational problem: 

%\begin{eqnarray}
%%& & \underset{p(\mobs|\mwst), p(\mact|\mobs)}{\operatorname{argmax}}{\mathbf{E}_{p(\mwst,\mact,\mobs)}[U(\mwst, \mact)] - \frac{1}{\beta_1}I(\mWst;\mObs) - \frac{1}{\beta_2}I(\mObs;\mAct) } \\ 
%& & \underset{p(\mobs|\mwst), p(\mact|\mobs)}{\operatorname{argmax}}{\mathbf{E}[U(\mwst, \mact)] - \frac{1}{\beta_1}I(\mWst;\mObs) - \frac{1}{\beta_2}I(\mObs;\mAct) } \\ 
%&=& \underset{p(\mobs|\mwst), p(\mact|\mobs)}{\operatorname{argmax}}{J(p(\mobs|\mwst), p(\mact|\mobs))} 
%\end{eqnarray}  

\begin{align}
 &\begin{aligned}
    &\underset{p(\mobs|\mwst), p(\mact|\mobs)}{\operatorname{argmax}}{\mathbf{E}_{p(\mwst,\mobs,\mact)}[U(\mwst, \mact)] - \frac{1}{\beta_1}I(\mWst;\mObs) } \\
    &\qquad  - \frac{1}{\beta_2}I(\mObs;\mAct) 
\end{aligned} \\
&= \underset{p(\mobs|\mwst), p(\mact|\mobs)}{\operatorname{argmax}}{J(p(\mobs|\mwst), p(\mact|\mobs))}  \label{eq:J}
\end{align}

Here we define $J$ as an overall objective function. Note that in this problem statement both the perceptual channel and the action channel are optimized. This is in contrast to traditional problem statements where a likelihood model is assumed to be given and the decision rule is built given this model. However, the coupling between action and perception falls out naturally by extending the rate distortion problem to Equation~\eqref{eq:J}.

Similar to the rate-distortion case of Equation~\eqref{eq:rd}, the solution is given by the following set of four analytic self-consistent equations \cite{Genewein2015}:

\begin{eqnarray}
p^*(\mobs|\mwst) &=&\frac{1}{Z(\mwst)}p(\mobs)\mathrm{exp}(\beta_1 \Delta F(\mwst,\mobs) ) \label{eq:sc1} \\
p(\mobs) &=&\sum\limits_{\mwst}p(\mwst)p^*(\mobs|\mwst) \label{eq:sc2}\\
p^*(\mact|\mobs) &=&\frac{1}{Z(\mobs)}p(\mact)\mathrm{exp}(\beta_2 \sum\limits_{\mwst}p(\mwst|\mobs)U(\mwst, \mact)) \label{eq:sc3}\\
p(\mact) &=&\sum\limits_{\mwst,\mobs}p(\mwst)p^*(\mobs|\mwst)p^*(\mact|\mobs) \label{eq:sc4}
\end{eqnarray} 
where $Z(\mwst)$ and $Z(x)$ denote the corresponding partition sums of world-state $\mwst$ and internal perception $\mobs$. The conditional probability $p(\mwst|x)$ is given by Bayes' rule
\[
p(\mwst|x) = \frac{p(\mwst)p^*(x|\mwst)}{p(x)}
\]
and
\[
\Delta F(\mwst,x):=\mathbf{E}_{p^*(a|x)}[U(\mwst,a)] -  \frac{1}{\beta_2} D_{\mathrm{KL}}(p^*(a|x)||p(a))
\] 
 is the free energy difference of the action stage. Equation (\ref{eq:sc1}) -- (\ref{eq:sc4}) can be computed by starting with arbitrary initial  distributions $q(\mobs)$, $q(\mact)$ and $q(\mact|\mobs)$ in lieu of $p(\mobs)$, $p(\mact)$ and $p(\mact|\mobs)$ and then iterating (\ref{eq:sc1}) to (\ref{eq:sc4}) until convergence. As the iterations involve evaluations of the utility function over all possible action and world-states, the computation can become very costly. Another drawback of such Blahut-Arimoto-style algorithms is that they cannot be applied straightforwardly to continuous problems, closed-form analytic solutions exist only for special cases. Here we propose an alternative online optimization method to solve this problem.
   
\section{Theoretical results}

%Biological agents---such as animals and humans---interact with a dynamic natural environment and constantly perceive rich sensory input to jointly adapt their sensory perception and action policy. 
In this study, we present an algorithm to update an agent's perceptual module and its behavioural policy---expressed through two separate parametric models---in a joint fashion under constrained information-processing resources according to the framework of information-theoretic bounded rationality.

\paragraph{Implementation of the perceptual channel $p(\mobs|\mwst)$}
   
\begin{figure}[tb]
      \centering
     \framebox{\parbox{.3\textwidth}{
     \centering
     \includegraphics{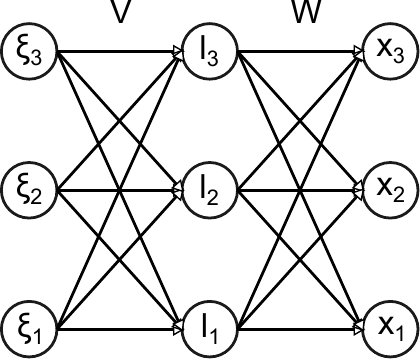}
     }}
      \caption{Illustration of the neural network model. The network possesses one input layer $\mathbf{\xi}$, one hidden layer $\mathbf{l}$ and one output layer $\mathbf{\mobs}$, and is parameterized by two real weight matrices $V$ and $W$. Note that it is not a prerequisite that each layer contains three neurons---the figure serves to illustrate the structure and the notation of the network, the number of neurons in each layer varies according to the problem setup. }
      \label{fig:NN}
   \end{figure}

We consider neural networks of the type depicted in Figure~\ref{fig:NN} as parameterized model to represent the perceptual distribution $p(\mobs|\mwst)$. The network possesses one input layer $\mathbf{\xi}$, one hidden layer $\mathbf{l}$ and one output layer $\mathbf{\mobs}$. $\mathbf{\xi}$ is a real-valued column vector representing the world-state $\mwst$. The synaptic weights between the input and the hidden layer are expressed as a real matrix $V$ and the weights between the hidden and the output layer are expressed as another real matrix $W$. The activation function in the hidden layer is a hyperbolic tangent function $ \phi(V^\dagger\mathbf{\xi}) = \mathrm{tanh}(V^\dagger\mathbf{\xi}) = 2 ./ (1 + \mathrm{exp}^{-2V^\dagger\mathbf{\xi}}) $. We apply a soft-max activation function in the output layer to compute the perceptual distribution with $p_{V,W}(x = x_i|\mathbf{\xi}) = \frac{\mathrm{exp }(\phi(\mathbf{\xi}^\dagger V)W{:, i})}{\sum\nolimits_{k=1}^{|X|}\mathrm{exp }(\phi(\mathbf{\xi}^\dagger V)W{:, k})}$. Accordingly, the gradients of the output distribution with respect to parameter matrices $V$ and $W$ are given by:
%$\frac{\partial}{\partial V_j} \mathrm{log}p_{v,w}(x_i|\mathbf{\xi})$ and $\frac{\partial}{\partial W_j} \mathrm{log}p_{v,w}(x_i| \mathbf{\xi})$

\begin{multline}
\frac{\partial}{\partial V}\mathrm{log}p_{V,W}(x_i|\mathbf{\xi}) = \mathbf{\xi}\cdot [ \phi'(\mathbf{\xi}^\dagger V) .\cdot W_{:, i}^\dagger] \\
-\sum\limits_{k=1}^{|X|}\left(p_{V,W}(x_k|\mathbf{\xi}) \cdot \mathbf{\xi}\cdot [ \phi'(\mathbf{\xi}^\dagger V) .\cdot W_{:, k}^\dagger]\right) 
\label{eq:dev_v}
\end{multline}

\vspace{-0.7cm}

 \begin{multline}
 \frac{\partial}{\partial W_j}\mathrm{log}p_{V,W}(x_i|\mathbf{\xi}) = \\
 \begin{cases}
 (1 -  p_{V,W}(x_j|\mathbf{\xi})) \cdot \phi(V^\dagger\mathbf{\xi}) & \quad \text{if } i = j \\
- p_{V,W}(x_j|\mathbf{\xi}) \cdot \phi(V^\dagger\mathbf{\xi})  & \quad \text{if } i\neq j 
\end{cases}
\label{eq:dev_w}
\end{multline}

A derivation of ($\ref{eq:dev_v}$) and ($\ref{eq:dev_w}$) can be found in the Appendix. 

%         \begin{figure*}[tb]
%      \centering
%     \framebox{\parbox{\textwidth}{
%     \centering
%     \includegraphics[width=\textwidth]{toy.pdf}
%     }}
%      \caption{Predator-prey example. (A) the predefined utility function $U(\mwst, \mact)$. (B) Development of the overall objective attained by the bounded-optimal actor over time. The red dashed line indicates the theoretic maximum value of the objective function computed by the Blahut-Arimoto algorithm. (C) The final strategy $p(\mact|\mwst)$ established by the actor after $10000$ iterations -- compare to Figure 6D in \cite{Genewein2015}.  }
%      \label{fig:toy}
%   \end{figure*}
   
\paragraph{Implementation of the action channel $p(\mact|\mobs)$}

Due to the abstract nature of the observation and action space, we assume here for simplicity discrete action choices. We therefore parameterize the action channel as a multinominal distribution: \\
  
\begin{equation}
\label{action_distribution}
p_{\mathbf{\eta}}(\mathbf{\mact}|\mobs) = \mathrm{exp } (\sum\limits_{i=1}^{n}\eta_i^{\mobs}\mact_i - \Psi(\mathbf{\eta^\mobs}) )
\end{equation}

with dimensionality $n = |\mAct| - 1$ and an auxiliary function $\Psi(\mathbf{\eta}) = \mathrm{log } (1 + \sum\nolimits_{i=1}^{n_\eta}\mathrm{exp}(\eta_i) )$. Note that the parameter $\eta$ is conditioned on observations $\mobs$. In our implementation $\eta$ is expressed and updated as a real-valued matrix with dimensionality $ n \times |\mObs|$.  We represent actions as a binary-valued vector in one-hot encoding having the form $\mathbf{\mact_i} = [\mact_0, \cdots, \mact_i, \cdots, \mact_n]$ with $\mact_i = 1$ and $\mact_j = 0$ for all $j \neq i$ where $0\leqslant  i \leqslant n$. The conventional constraint of a conditional distribution $\sum\limits_{\mathbf{\mact}}p_{\mathbf{\eta}}(\mathbf{\mact}|\mobs) = 1$  is satisfied by defining $p_{\mathbf{\eta}}(\mathbf{\mact_0}|\mobs) = 1 - \sum\nolimits_{i=1}^{n}p_{\mathbf{\eta}}(\mathbf{\mact_i}|\mobs)$ . Thus, the gradient of the action distribution with respect to the parameter $\eta$ is given by

\begin{equation}
\frac{\partial}{\partial \eta} \mathrm{log}p_{\mathbf{\eta}}(\mact|\mobs) = \begin{bmatrix}
\frac{\partial}{\partial \eta_1} \mathrm{log}p_{\mathbf{\eta}}(\mact|\mobs) \\
\vdots \\
\frac{\partial}{\partial \eta_{n}} \mathrm{log}p_{\mathbf{\eta}}(\mact|\mobs)
\end{bmatrix} = \begin{bmatrix}
\mact_1 - \frac{\mathrm{exp} (\eta_1^\mobs)}{1 + \sum_i\mathrm{exp}(\eta_i^\mobs)} \\
\vdots \\
\mact_n - \frac{\mathrm{exp} (\eta_n^\mobs)}{1 + \sum_i\mathrm{exp}(\eta_i^\mobs)} 
\end{bmatrix} 
\label{eq:eta}
\end{equation}

\paragraph{Parameter updates} \label{subsec:update}
In the course of the simulation, the bounded rational decision-maker constantly updates the parameters representing the perceptual and the action channel. The overall objective in (\ref{eq:J}) is expressed as a parametric function of $V$, $W$ and $\eta$:

\begin{multline}
J = \sum_{\mwst} \sum_{\mobs} \sum_{\mact} p(\mwst) p(\mobs|\mwst) p(\mact|\mobs) \\
 \cdot \left( U(\mwst, \mact) - \frac{1}{\beta_1} \mathrm{log} \frac{p_{V,W}(\mobs|\mwst)}{p_{v,w}(\mobs)} - \frac{1}{\beta_2} \mathrm{log} \frac{p_{\mathbf{\eta}}(\mact|\mobs)}{p_{V,W,{\mathbf{\eta}}}(\mact)} \right) 
\end{multline}

By defining an auxiliary term $j(\mwst, \mobs, \mact) = U(\mwst, \mact) -  \frac{1}{\beta_1} \mathrm{log} \frac{p_{V,W}(\mobs|\mwst)}{p_{V,W}(\mobs)} - \frac{1}{\beta_2} \mathrm{log} \frac{p_{\mathbf{\eta}}(\mact|\mobs)}{p_{V,W,{\mathbf{\eta}}}(\mact)} $ the objective can be rewritten as $J = \sum_{\mwst, \mobs, \mact} p(\mwst, \mobs, \mact) j(\mwst, \mobs, \mact)$. 
Here we apply the log-trick to transform the derivative of $J$ into an expected value by noticing that for any parametric function $f_\theta(x)$ the equation $\sum_x[\frac{\partial}{\partial \theta}p_\theta(x)]f_\theta(x) = \sum_x p_\theta(x)[\frac{\partial}{\partial \theta}\mathrm{log}p_\theta(x)]f_\theta(x)$ is valid. This trick allows us to rewrite the derivative of the overall objective as follows:

 \begin{eqnarray}
\frac{\partial J}{\partial V}&=&\left< \left( \frac{\partial}{\partial V} \mathrm{log}p_{V,W}(\mobs|\mwst) \right)j(\mwst, \mobs, \mact) \right>_{\mwst,\mobs,\mact} \\
\frac{\partial J}{\partial W}&=&\left< \left( \frac{\partial}{\partial W} \mathrm{log}p_{V,W}(\mobs|\mwst) \right)j(\mwst, \mobs, \mact) \right>_{\mwst,\mobs,\mact} \\
\frac{\partial J}{\partial \eta}&=&\left< \left( \frac{\partial}{\partial \eta} \mathrm{log}p_{\mathbf{\eta}}(\mact|\mobs) \right)j(\mwst, \mobs, \mact) \right>_{\mwst,\mobs,\mact}
 \end{eqnarray}

The expectation value can be approximated by drawing $N$ sample triplets $\{\mwst,\mobs,\mact\}$ from the joint distribution $p_{V,W,{\mathbf{\eta}}}(\mwst,\mobs,\mact)$. The number of samples $N$ governs the accuracy of the approximation. A large $N$ provides high accuracy but demands vast computational resources. Setting the batch size to $1$ economizes the computational cost at every iteration by avoiding the expensive evaluation of the summand function, thus leading to an effective online rule for parameter updates as is done in stochastic gradient ascent---see \cite{Felix2016} for a similar method. We apply a soft update rule to optimize parameters in an online fashion by introducing the learning rate $\alpha > 0$

\begin{equation}
\theta \leftarrow \theta + \alpha \cdot \frac{\partial J}{\partial \theta} \label{eq:lr} 
\end{equation}
for each parameter $\theta \in \{V, W, {\mathbf{\eta}}\}$. Note that stochastic gradient ascent does not always converge to a global maximum. The global maximum is assumed only when the objective function is concave and the learning rates decrease with an appropriate rate, otherwise a local maximum might be attained \cite{Bottou1998}. Therefore, the online rule for the problem described is not guaranteed to converge to global optimal solutions and should be treated carefully by using small learning rates $\alpha$.

\section{Experimental Results}
\label{sec:exp}
Iterating the analytical solutions (Equations \eqref{eq:sc1}-\eqref{eq:sc4}) requires the evaluation of the utility function $U(\mwst,\mact)$ for \emph{all} $\mwst,\mact$ pairs in each iteration step. A major advantage of the gradient-update scheme derived in the previous section is that it is suitable for on-line updates, i.e. one iteration can be performed after every interaction of the robot with the environment. In such a scheme world-states $\mwst$ are generated (randomly) by the environment $\hat{\mwst} \sim p(\mwst)$. Each $\hat{\mwst}$ is processed by the perceptual stage of the agent, which samples a percept $\hat{\mobs} \sim p_{V,W}(\mobs|\hat{\mwst})$ (in our case, drawing a sample from the distribution that results from the softmax output of the neural network). Similarly, the agent samples an action $\hat{\mact} \sim p_{\mathbf{\eta}}(\mact|\hat{\mobs})$. This leads to a roll-out which allows to evaluate $U(\hat{\mwst},\hat{\mact})$, and to perform one stochastic gradient update step. In the following, we first compare this on-line stochastic gradient update scheme against solutions obtained from iterating the analytical solution equations. Afterwards we demonstrate the scheme on a more challenging task in a simulated robot environment.

\begin{figure*}[htb]
      \centering
     \framebox{\parbox{.97\textwidth}{
     \centering
     \includegraphics[width=.97\textwidth]{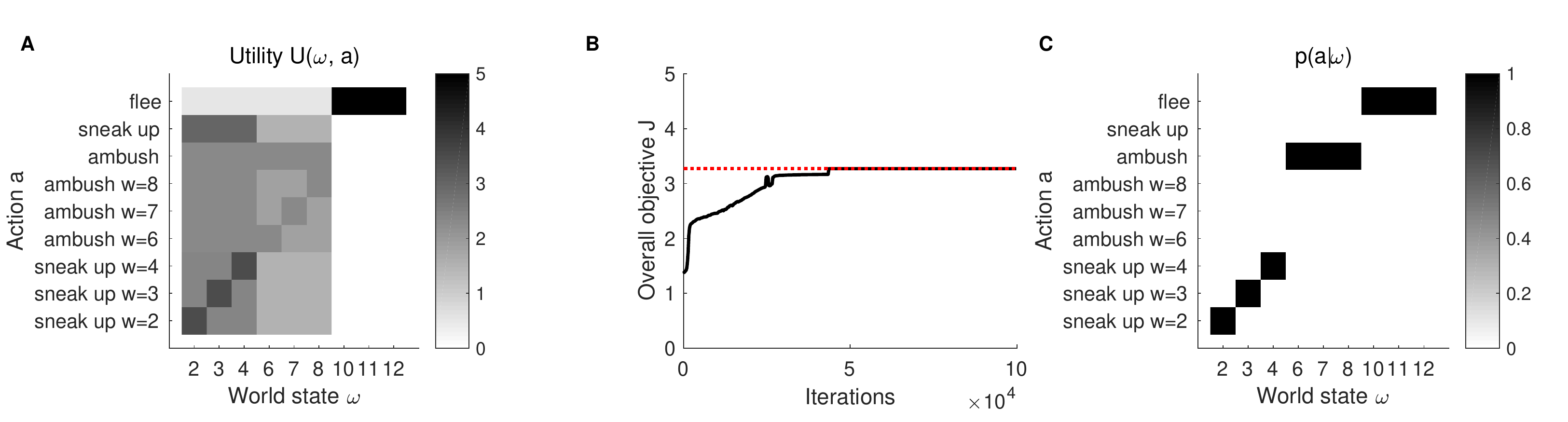}
     }}
      \caption{Predator-prey example with a neural network for the perceptual stage and a multinomial distribution for the action-stage. (A) Utility function $U(\mwst, \mact)$, see \cite{Genewein2015} for a detailed description. (B) Black line: evolution of the  objective (Equation \eqref{eq:J}) during gradient-update iterations (Equation \eqref{eq:lr}). Red dashed line: baseline objective-value achieved by iterating the analytical solutions (Equations \eqref{eq:sc1}-\eqref{eq:sc4}) until convergence. (C) Final ``behavior'' $p(\mact|\mwst)$ after $100000$ iterations -- compare Figure 6D in \cite{Genewein2015}.}
      \label{fig:toy}
\end{figure*}

\subsection{Comparison with baseline}
To empirically verify the convergence of our gradient update scheme and the correctness of the resulting solution, we compare against the (analytical) solution obtained by iterating the set of self-consistent equations as given in \cite{Genewein2015}. To this end, we use the ``predator-prey'' example from \cite{Genewein2015}. In the example, a fictional animal encounters other animals, which can either be prey that should be hunted, or predators that should be avoided. To decide which action $\mact$ to take, the animal has a perceptual sensor to determine the size $\mwst$ of the encountered animal. The example is described by the utility function $U(\mwst,\mact)$ shown in Figure \ref{fig:toy}A. Animals belong to one of three groups: small, medium-sized and large animals. All large animals are predators, thus the only action that yields non-zero utility is ``flee'' (regardless of the particular size of the large animal). For each of the small animals, a specific hunting-action yields the highest utility, therefore it is relevant to distinguish between the individual animals of the small-group. In contrast, for the animals of the medium-sized group the specific hunting-actions yield the same utility as a generic hunting-action that works equally well for all medium-sized animals. The example clearly illustrates the importance of coupling perception with the downstream action-part of the agent. The distinction between the individual animals of the medium- and large-sized groups is irrelevant for acting. Thus, spending (computational) capacity on the perceptual channel for this distinction is lavish and should be avoided, particularly if the capacity of the perceptual channel is limited.

The original example in \cite{Genewein2015} used categorical distributions for perception $p(\mobs|\mwst)$ and action $p(\mact|\mobs)$. Here, we use a neural network with one hidden layer for perception $p_{V,W}(\mobs|\xi)$ (with $\xi$ being a binary encoding of $\mwst$) and a multinomial distribution (parameterized as given by Equation \eqref{action_distribution}) for the action-stage $p_{\mathbf{\eta}}(\mathbf{\mact}|\mobs)$ (with $\mathbf{\mact}$ being a one-hot encoding of the action $\mact$). The neural network consisted of four input neurons, 20 hidden and 13 output neurons, initialized with Glorot's scheme (also known as Xavier-initialization) \cite{Glorot10}. The parameters $\eta$ were initialized such that all actions were equally probable. 
We found that convergence of the gradient-update scheme crucially depends on using different learning rates for the perceptual channel ($V,W$) and the action channel (${\mathbf{\eta}}$). Figure \ref{fig:toy}B shows the evolution of the objective value (Equation \eqref{eq:J}) during gradient-update iterations of $V,W,{\mathbf{\eta}}$ (Equation \eqref{eq:lr}) using $\beta_1=8$ (corresponding to large perceptual capacity) and $\beta_2=10$ (high-capacity action channel). We used a learning rate of $\alpha_{V,W}=0.006$ for the perceptual channel and $\alpha_{\mathbf{\eta}}=0.014$ for the action channel.
The dashed red line in the panel indicates the baseline, that is the value of the objective function obtained by iterating the set of analytical solutions  (Equations \eqref{eq:sc1}-\eqref{eq:sc4}) until convergence as in \cite{Genewein2015}. As shown in the figure, the gradient update scheme converges to a solution with the same objective-value as the analytical baseline method (after roughly 50000 iterations). Panel C of Figure \ref{fig:toy} shows the corresponding behavior $p_{V,W,{\mathbf{\eta}}}(\mact|\mwst)=\sum_\mobs p_{V,W}(\mobs|\mwst) p_{\mathbf{\eta}}(\mact|\mobs)$ (omitting binary/one-hot encoding from the notation for simplicity) after $100000$ gradient-update iterations. Comparing panel C against Figure 6D in \cite{Genewein2015} shows, that the solution obtained from the gradient update scheme is qualitatively identical to the solution obtained by iterating the set of self-consistent equations. Importantly, the solution reflects the intuition, that an information-optimal agent does not distinguish between the individual animals of the medium and large groups, even if the computational capacity of the perceptual channel would in principle allow for such a distinction.

%\textcolor{red}{
%To support our story we should ideally show here the perceptual channel $p(x|w)$ as well and also show how it changes under limited resources, i.e. show the solution for low $\beta_1$ as well. On top of that it would perhaps also be supportive for our story to show the comparison against a hand-crafted perceptual model (i.e. handcrafted likelihood). But then we would in essence show (almost) the whole example from the Frontiers paper again. If we have space, we should think about that.
%}

We have performed the same comparison for the other settings of $\beta_1, \beta_2$ in \cite{Genewein2015}, corresponding to either low capacity of the perceptual- or the action-channel, and found that the gradient scheme converges to solutions that are qualitatively identical (same objective value, same behavior $p(a|w)$) to solutions obtained from iterating the self-consistent equations. We conclude that the gradient update scheme in conjunction with a neural network for the perceptual stage and a multinomial distribution for the action-stage successfully matches the analytical baseline. Due to the limitations in length of the manuscript, we have omitted further plots of the empirical baseline comparison.

\subsection{Robot Simulation}
\begin{figure}[htb]
     \centering
     \framebox{\parbox{2.2in}{
     \centering
     \includegraphics[width=2.2in]{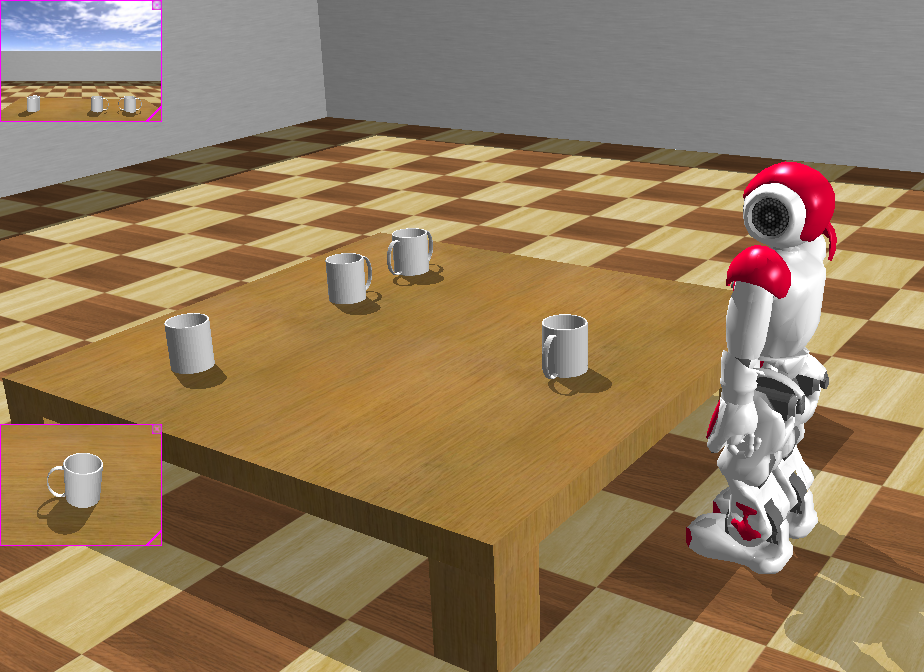}
     }}
      \caption{Task setup in Nao simulation environment.}
      \label{fig:nao}
\end{figure}

\begin{figure}[hbt]
      \centering
     \framebox{\parbox{3.3in}{
     \centering
     \includegraphics[width=3.3in]{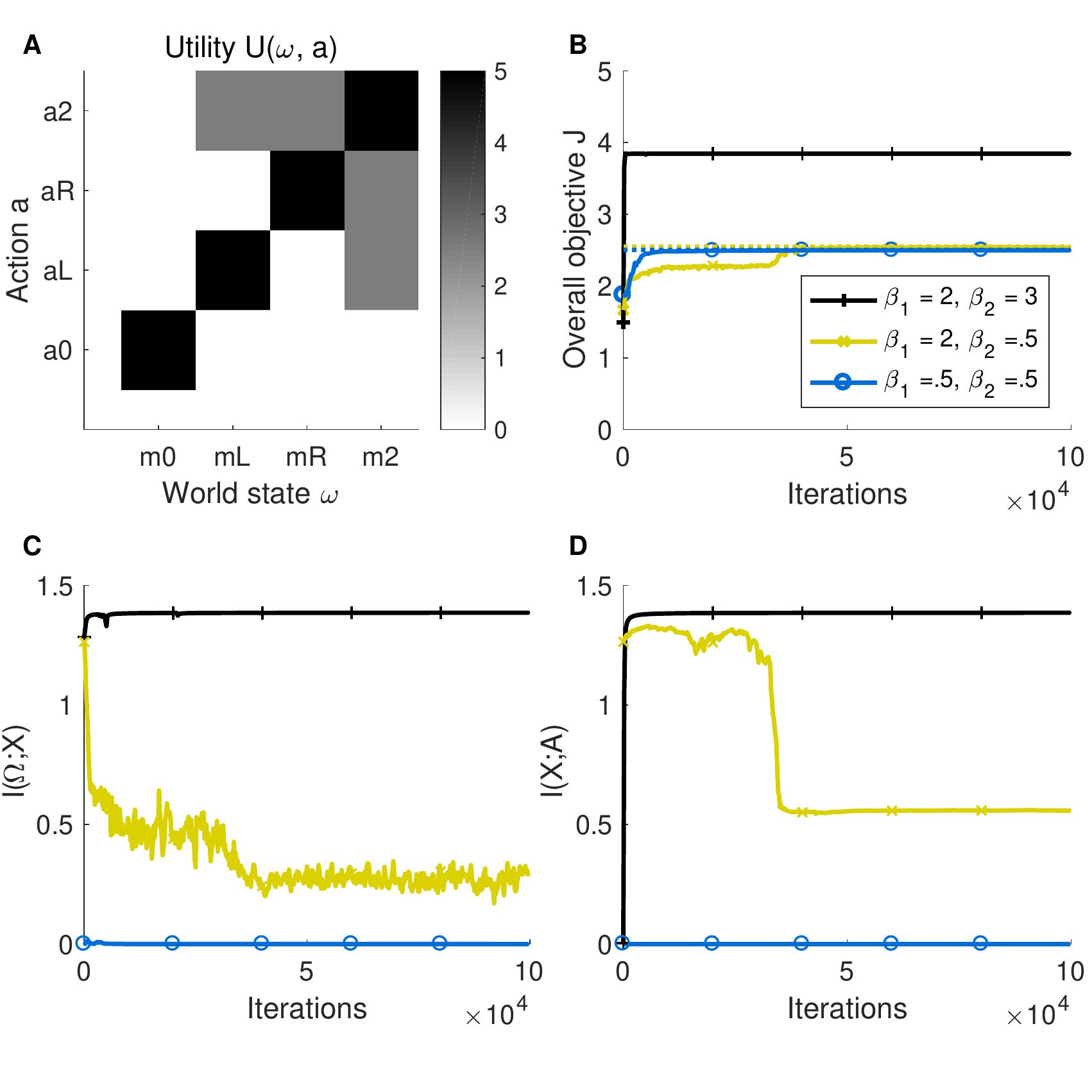}
     }}
      \caption{(A) The utility function $U(\mwst, \mact)$. $m0$ indicates the mug without any handle, $mL$ and $mR$ represent the mug with one handle (rotated correspondingly) and $m2$ symbolizes the mug having handles on both sides. For each mug there is one ``most suitable'' action that yields the highest utility (in case of no handles, executing no lift is defined to be optimal). Additionally, lifting the one-handle cup with both hands leads to a non-zero utility, corresponding to a successful lift, but using more effort than necessary. (B) Evolution of the objective-value during on-line gradient-update training. Three different computational limitations are compared here, dashed lines: baseline objective-value achieved by iterating the analytical solutions. (C) \& (D) Evolution of the mutual information of the perceptual channel $I(\mWst;\mObs)$ and the action channel $I(\mObs;\mAct)$.
      }
      \label{fig:nao_dev}
\end{figure}

We also test our method in a simulated robotic environment to illustrate the usage of the proposed update principle for sensorimotor coupling with parametric perception and action modules. To this end, we designed a simplified grasping task with a simulated Nao robot. In the simulation, the robot is positioned next to a table with $4$ mugs on it (see Figure \ref{fig:nao}). The mugs differ in the number and orientation of the handles. One mug (m0) has no handle at all, two mugs have one handle (positioned such that the handle is either to the left (mL) or to the right (mR)) and one mug has handles on both sides (m2), allowing a direct grasp with either the left or the right hand of the robot. The Nao robot has two cameras---in this simulation we make use of the chest-camera which shows the area on the table directly in front of the robot (see the bottom left inlet in Figure \ref{fig:nao}). Based on this camera input, the (bounded-rational) agent has to decide how to grasp the mug appropriately. We defined $4$ possible actions: lift the mug with both hands (a2), lift the mug with the left hand (aL), lift with the right hand (aR) or execute no lift (a0). Additionally, we have defined the following utility function shown in Figure~\ref{fig:nao_dev}A, where each mug has one preferred action (yielding the highest utility) and the case of grasping a single-handle mug with both hands, which yields slightly lower utility.

As in the previous example, the perceptual stage of the robot is implemented by a neural network with one hidden layer (192 input neurons, four hidden and four output neurons, Xavier-initialization), and the action stage is implemented with a multinomial distribution (parameterized according to Equation \eqref{action_distribution}, initialized to have uniform probabilities over actions). The perceptual and action stage are then learned through interaction with the environment, using the stochastic gradient update scheme proposed in this paper. At the beginning of each trial, one mug is randomly selected according to $p(\mwst)$ (uniform distribution in our case) and placed in front of the agent. A $16\times12$ image from the chest camera of the robot is then fed into the neural network for perceptual processing (2D-image is flattened into 1D vector). Accordingly, an observation $\hat{\mobs}$ and an action $\hat{\mact}$ are sampled by the agent. After evaluating the utility $U(\hat{\mwst}, \hat{\mact})$, the model parameters $V$, $W$ and ${\mathbf{\eta}}$ are updated with a gradient ascent step as described in the previous section.

Figure~\ref{fig:nao_dev} shows three experiments with different computational capacity of the perception- and action-channels, they are: high-capacity channels ($\beta_1 = 2$ and $\beta_2 =3$, $\alpha_{V,W}=0.035$, $\alpha_{\mathbf{\eta}}=0.7$), high perceptual capacity combined with low action capacity ($\beta_1 = 2$ and $\beta_2 =0.5$, $\alpha_{V,W}=0.001$, $\alpha_{\mathbf{\eta}}=0.34$), and low-capacity channels ($\beta_1 = \beta_2 = 0.5$, $\alpha_{V,W}=0.004$, $\alpha_{\mathbf{\eta}}=0.028$). Panel B shows how the objective value evolves during training. In all cases, the on-line update scheme converges to a stable solution. The dashed lines show the optimal solution. Panel C and D show evolution of the mutual information of the perceptual channel $I(\mWst;\mObs)$ and the action channel $I(\mObs;\mAct)$. The mutual information on the perceptual and action stage is high with high computational capacity. Lowering the capacity of the action channel leads to a reduction of mutual information of both channels. Note that for the second case only the information processing price on the action stage is changed, the perceptual channel adjusts accordingly. Under (very) low computational capacity, the agent develops a single action strategy such that it requires no computation at all (channel capacity for both perception and action is effectively zero). 

\begin{figure}[hbt]
      \centering
     \framebox{\parbox{3.3in}{
     \centering
     \includegraphics[width=3.3in]{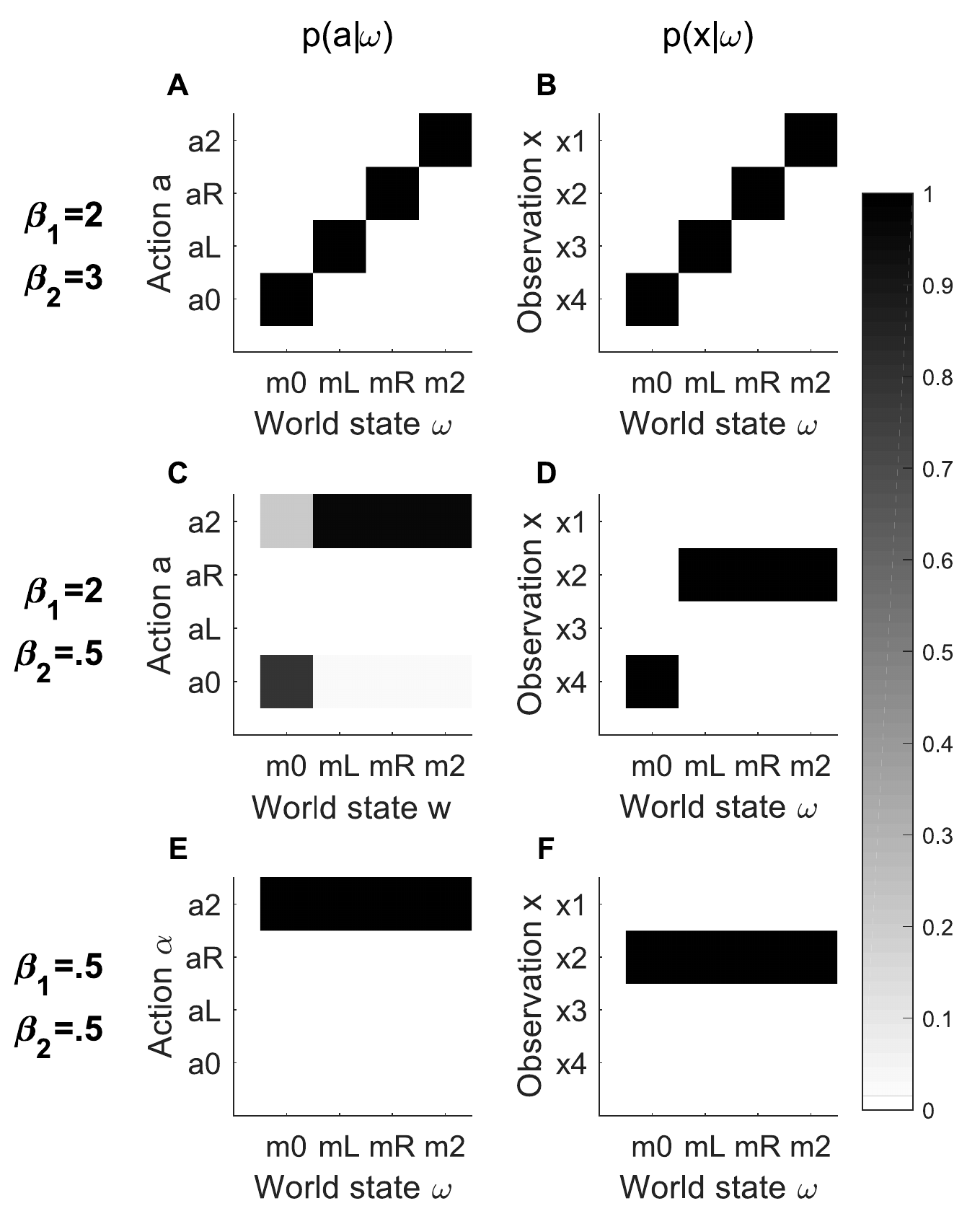}
     }}
      \caption{Comparing the final behaviour of a bounded-optimal agent after $100 000$ iterations under different computational constrains. (A) \& (B) results of the simulation with high computational capacity ($\beta_1 = 2, \beta_2 = 3$). (C) \& (D) results with low action channel capacity ($\beta_1 = 2, \beta_2 = 0.5$). (E) \& (F) results with (very) low computational capacity ($\beta_1 = \beta_2 = 0.5$). }
      \label{fig:nao_beh}
\end{figure}

Figure~\ref{fig:nao_beh} shows the behaviour of the robot after convergence.With high computational capacity, the robot learns to associate the camera images with the best possible action for each mug (Panel A \& B). If the action channel does not have sufficient capacity, the agent is not able to apply specific actions to specific contexts, therefore, its policy collapses  into two modes: lift with both hands or do nothing at all (Panel C). Accordingly, the agent spends less information processing in the perceptual channel such that it only discriminates between mugs with handle(s) and mugs without handle (Panel D). Under (very) low computational capacity, the agent always chooses to lift the mug with both hands which requires no computation. (Panel E $\&$ F)

%Figure \ref{fig:nao_res} shows two experiments with different computational capacity of the perception- and action-channels. High-capacity channels ($\beta_1 = \beta_2 = 100$, $\alpha_{V,W}=0.0007$, $\alpha_{\mathbf{\eta}}=1.2$) are shown in the upper half of the figure (panels A to D)---the low-capacity experiment ($\beta_1 = \beta_2 = 0.5$, $\alpha_{V,W}=0.004$, $\alpha_{\mathbf{\eta}}=0.028$) is shown in panels E to H. The left column (panels A and E) show how the objective value evolves during training. In both cases, the on-line update scheme converges to a stable solution. The dashed red lines show the optimal solution.
%% value obtained by iterating the analytical solutions as a comparison (however, the analytical solutions cannot use the high-dimensional image input but use a low-dimensional representation of the world state $\mwst \in \{\text{M0, M1L, M1R, M2}\}$).
%Panels B and F show the ``behaviour'' of the robot after convergence. With high computational capacity, the robot learns to associate the camera images with the best possible action for each mug. Correspondingly, the mutual information on the perceptual and action stage is relatively high as shown by panels C and D. In contrast, under (very) low computational capacity, the agent always chooses to lift the mug with both hands. Panels G and H reveal that such a strategy requires no computation at all (channel capacity for both perception and action is effectively zero). Under this constraint, always picking to lift with both hands is the best possible strategy.

\section{Discussion}
                      
In this study we propose a novel online optimization rule to find bounded-optimal perception-action coupling in serial perception-action systems. The perceptual channel is implemented as a multi-layer neural network while the action channel is represented by a parametric distribution, which was a multinomial in our case. Our method is illustrated with a NAO robot simulator.

% blahut-arimoto
The proposed algorithm can be improved in several ways.
In the case of rate distortion equation~\eqref{eq:rd}, the Blahut-Arimoto algorithm is guaranteed to converge to a unique maximum [see Section 2.1.1 in \cite{Tishby1999}].  When we consider an extension of bounded rational problems to systems with multiple information-processing stages, since there is no convergence-proof, it cannot be ruled out that the solutions obtained by iterating the self-consistent equations until numerical convergence are only local optima. One future improvement would therefore be to better theoretically understand the convergence properties of the extended rate distortion problem. 

%The one with maximal objective value is then selected as estimated optimal solutions and presented as red dashed lines in Fig~\ref{fig:toy} and Fig~\ref{fig:nao_res}. 

% learning rate
Another issue is that the learning rates $\alpha$ for the parameter updates (see equation \eqref{eq:lr}) have a significant impact on the development of the bounded-optimal behaviors. Inappropriate learning rates easily lead to an optimization failure in that the bounded-rational decision-maker is unable to find the bounded-optimal solutions. We choose a grid-search method to optimize the learning rates.  A future improvement would require a better understanding of the relationship between the learning rates in the perceptual and action module and to study better optimization procedures for choosing the learning rates accordingly.

In information-theoretic bounded rationality, the main assumption is that the decision-maker's behavioral policy may not deviate too much from some prior policy. Deviations from the prior policy are costly and modeled through the Kullback-Leibler divergence (Section 1.A). This is similar to other proposed regularization techniques from robotics such as Trust Region Policy Optimization (TRPO, \cite{Schulman2015}) and Relative Entropy Policy Search (REPS, \cite{Peters2010}). The main difference lies in the choice of the prior behavioral policy. 
In both, TRPO and REPS, the prior policy represents the agent's behavior at a previous iteration of the optimization procedure or a set of initial expert trajectories (in imitation-learning). In our approach, on the other hand, the agent's prior policy is optimal w.r.t. the information-theoretic constraints (Eq. (3)), encouraging the agent to ignore irrelevant sensory information that has little impact on the reward.

%In both, TRPO and REPS, the prior policy represents the agent's behavior at a previous iteration of the optimization procedure. In our approach on the other hand, the agent's prior policy is optimal under information-theoretic aspects encouraging to neglect such information on the perceptual side that has little impact on the reward.

Conceptually, bounded rationality has its most obvious application when the information-processing capacity is limited by physical constraints like time or space constraints. In our present model this was not the case, which raises the question why one would restrict the system to a smaller capacity than it might naturally have. Apart from possible effects on learning speed that we did not investigate here, restricting the channel capacity creates a bottleneck that filters out relevant information and creates abstractions that are useful for generalization \cite{Tishby1999}. For instance in panel D of figure~\ref{fig:nao_beh}, two abstract percepts emerge: ``mug with handle(s)" (x2) and ``mug without handles" (x4).

In summary, our information-theoretic principle~\eqref{eq:J} for perception-action coupling provides a novel generic  principled method and could in principle be applied to combine any parameterized perception and action modules. Compared to the existing literature, this approach is most similar in spirit to approaches that learn particular perceptual features that are most useful to solve a particular task \cite{Agrawal2016, Jonschkowski2015, Levine 2016, Piater2011}. Here this feature search is integrated in a single bounded rational optimization problem.

% limitation of continue case
%Following the work of \cite{Genewein2015}, we have extended the application regime of the bounded-optimal decision-making from the pure abstract perception into a continuous space. However, the intermediate observation and action spaces stay abstract. Out further research will extend the both channels into continuous spaces, in order to extend the application possibility in robotics of our method. 

%\section*{References}
%\bibliography{library}

%%%%%%%%%%%%%%%%%%%%%%%%%%%%%%%%%%%%%%%%%%%%%%%%%%%%%%%%%%%%%%%%%%%%%%%%%%%%%%%%
\section*{APPENDIX} \label{App:AppendixA}

\subsection{Partial derivative of the perceptual distribution with respect to $V$} 
\vspace{-0.45cm}
\begin{align*}
  &\frac{\partial}{\partial V}\mathrm{log}p_{V,W}(x_i|\mathbf{\xi}) = 
\frac{\partial}{\partial V}\mathrm{log}\frac{\mathrm{exp }(\phi(\mathbf{\xi}^\dagger V)W_{:, i})}{\sum\limits_{k=1}^{|X|}\mathrm{exp }(\phi(\mathbf{\xi}^\dagger V)W_{:, k})}  \\  
&= \frac{\partial}{\partial V}(\phi(\mathbf{\xi}^\dagger V)W_{:, i})) - 
\frac{\partial}{\partial V} \mathrm{log} \sum\limits_{k=1}^{|X|}\mathrm{exp }(\phi(\mathbf{\xi}^\dagger V)W_{:, k}) \\
&\begin{aligned}
    &=  \frac{\partial}{\partial V}(\mathbf{\xi}^\dagger V) \cdot [ \phi'(\mathbf{\xi}^\dagger V) .\cdot W_{:, i}^\dagger] \\
    &\qquad - \frac{\sum\limits_{k=1}^{|X|}\mathrm{exp }(\phi(\mathbf{\xi}^\dagger V)W_{:, k}) \cdot \frac{\partial}{\partial V}(\phi(\mathbf{\xi}^\dagger V)W_{:, k})}{\sum\limits_{k=1}^{|X|}\mathrm{exp }(\phi(\mathbf{\xi}^\dagger V)W_{:, k})}
\end{aligned} \\
&\begin{aligned}
    &=  \mathbf{\xi}\cdot [ \phi'(\mathbf{\xi}^\dagger V) .\cdot W_{:, i}^\dagger]\\
    &\qquad - \sum\limits_{k=1}^{|X|}\left(p_{V,W}(x_k|\mathbf{\xi}) \cdot \mathbf{\xi}\cdot [ \phi'(\mathbf{\xi}^\dagger V) .\cdot W_{:, k}^\dagger]\right) 
\end{aligned}
\end{align*}

%\begin{eqnarray}
%& &\frac{\partial}{\partial V}\mathrm{log}p_{v,w}(x_i|\mathbf{\xi}) = 
%\frac{\partial}{\partial V}\mathrm{log}\frac{\mathrm{exp }(\phi(\mathbf{\xi}^\dagger V)W_{:, i})}{\sum\limits_{k=1}^{|X|}\mathrm{exp }(\phi(\mathbf{\xi}^\dagger V)W_{:, k})}  \\
%&=& \frac{\partial}{\partial V}(\phi(\mathbf{\xi}^\dagger V)W_{:, i})) - 
%\frac{\partial}{\partial V} \mathrm{log} \sum\limits_{k=1}^{|X|}\mathrm{exp }(\phi(\mathbf{\xi}^\dagger V)W_{:, k}) \\
%&=&  \frac{\partial}{\partial V}(\mathbf{\xi}^\dagger V) \cdot [ \phi'(\mathbf{\xi}^\dagger V) .\cdot W_{:, i}^\dagger] - \frac{\sum\limits_{k=1}^{|X|}\mathrm{exp }(\phi(\mathbf{\xi}^\dagger V)W_{:, k}) \cdot \frac{\partial}{\partial V}(\phi(\mathbf{\xi}^\dagger V)W_{:, k})}{\sum\limits_{k=1}^{|X|}\mathrm{exp }(\phi(\mathbf{\xi}^\dagger V)W_{:, k})}\\
%&=&\mathbf{\xi}\cdot [ \phi'(\mathbf{\xi}^\dagger V) .\cdot W_{:, i}^\dagger] - \sum\limits_{k=1}^{|X|}\left(p_{v,w}(x_k|\mathbf{\xi}) \cdot \mathbf{\xi}\cdot [ \phi'(\mathbf{\xi}^\dagger V) .\cdot W_{:, k}^\dagger]\right) 
%\end{eqnarray}

\subsection{Partial derivative of the perceptual distribution with respect to $W$} 
\paragraph{$j = i$}

\begin{align*}
  &\frac{\partial}{\partial W_i}\mathrm{log}p_{V,W}(x_i|\mathbf{\xi}) \\
  &= \frac{\partial}{\partial W_i}\mathrm{log}\frac{\mathrm{exp }(\phi(\mathbf{\xi}^\dagger V)W_{:, i})}{\sum\limits_{k=1}^{|X|}\mathrm{exp }(\phi(\mathbf{\xi}^\dagger V)W_{:, k})}  \\
 &\begin{aligned}
    &=  \frac{\partial}{\partial W_i}(\phi(\mathbf{\xi}^\dagger V)W_{:, i})) \\
    &\qquad - \frac{\partial}{\partial W_i} \mathrm{log} \sum\limits_{k=1}^{|X|}\mathrm{exp }(\phi(\mathbf{\xi}^\dagger V)W_{:, k})
\end{aligned} \\
&= \phi(V^\dagger\mathbf{\xi})  - \frac{\mathrm{exp }(\phi(\mathbf{\xi}^\dagger V)W_{:, i})}{\sum\limits_{k=1}^{|X|}\mathrm{exp }(\phi(\mathbf{\xi}^\dagger V)W_{:, k})} \cdot\phi(V^\dagger\mathbf{\xi}) \\
&= (1 -  p_{V,W}(x_i|\mathbf{\xi})) \cdot \phi(V^\dagger\mathbf{\xi})
\end{align*}

% \begin{eqnarray}
%& &\frac{\partial}{\partial W_i}\mathrm{log}p_{v,w}(x_i|\mathbf{\xi}) = 
%\frac{\partial}{\partial W_i}\mathrm{log}\frac{\mathrm{exp }(\phi(\mathbf{\xi}^\dagger V)W_{:, i})}{\sum\limits_{k=1}^{|X|}\mathrm{exp }(\phi(\mathbf{\xi}^\dagger V)W_{:, k})}  \\
%&=& \frac{\partial}{\partial W_i}(\phi(\mathbf{\xi}^\dagger V)W_{:, i})) - 
%\frac{\partial}{\partial W_i} \mathrm{log} \sum\limits_{k=1}^{|X|}\mathrm{exp }(\phi(\mathbf{\xi}^\dagger V)W_{:, k}) \\
%&=& \phi(V^\dagger\mathbf{\xi})  - \frac{\mathrm{exp }(\phi(\mathbf{\xi}^\dagger V)W_{:, i})}{\sum\limits_{k=1}^{|X|}\mathrm{exp }(\phi(\mathbf{\xi}^\dagger V)W_{:, k})} \cdot\phi(V^\dagger\mathbf{\xi}) \\
%&=& (1 -  p_{v,w}(x_i|\mathbf{\xi})) \cdot \phi(V^\dagger\mathbf{\xi})
%\end{eqnarray}

\paragraph{$j \neq i$}

\begin{align*}
  &\frac{\partial}{\partial W_j}\mathrm{log}p_{V,W}(x_i|\mathbf{\xi}) \\
  &= \frac{\partial}{\partial W_j}\mathrm{log}\frac{\mathrm{exp }(\phi(\mathbf{\xi}^\dagger V)W_{:, i})}{\sum\limits_{k=1}^{|X|}\mathrm{exp }(\phi(\mathbf{\xi}^\dagger V)W_{:, k})}  \\
 &\begin{aligned}
    &=  \frac{\partial}{\partial W_j}(\phi(\mathbf{\xi}^\dagger V)W_{:, i})) \\
    &\qquad - \frac{\partial}{\partial W_j} \mathrm{log} \sum\limits_{k=1}^{|X|}\mathrm{exp }(\phi(\mathbf{\xi}^\dagger V)W_{:, k})
\end{aligned} \\
&= - \frac{\mathrm{exp }(\phi(\mathbf{\xi}^\dagger V)W_{:, j})}{\sum\limits_{k=1}^{|X|}\mathrm{exp }(\phi(\mathbf{\xi}^\dagger V)W_{:, k})} \cdot\phi(V^\dagger\mathbf{\xi}) \\
&= - p_{V,W}(x_j|\mathbf{\xi}) \cdot \phi(V^\dagger\mathbf{\xi})
\end{align*}

\end{document}